\documentclass{article}
\usepackage{PRIMEarxiv}

\usepackage[utf8]{inputenc} 
\usepackage[T1]{fontenc}    
\usepackage{hyperref}       
\usepackage{url}            
\usepackage{booktabs}       
\usepackage{amsfonts}       
\usepackage{nicefrac}       
\usepackage{microtype}      
\usepackage{lipsum}
\usepackage{fancyhdr}       
\usepackage{graphicx}       
\graphicspath{{media/}}     
\usepackage{amsmath} 
\usepackage[ruled,vlined,linesnumbered]{algorithm2e}
\usepackage[
    backend=biber,
    style=authoryear,
    natbib=true,
    maxcitenames=1,
    maxbibnames=5,
    url=false,
    doi=true,
    eprint=false
]{biblatex}
\usepackage{xcolor}
\usepackage{multirow,colortbl,graphicx,makecell,adjustbox}
\usepackage{chngcntr}
\usepackage[section]{placeins}

\title{Goal-Directed Search Outperforms Goal-Agnostic Memory Compression in Long-Context Memory Tasks}

\author{
  Yicong Zheng, Kevin L. McKee, Thomas Miconi, Zacharie Bugaud, Mick van Gelderen, Jed McCaleb \\
  Astera Institute \\
  Emeryville, CA, USA\\
  \texttt{\{alan, kevin, thomasmiconi, zacharie, mick, jed\}@astera.org} \\
}

\begin{document}
\maketitle

\begin{abstract}

How to enable human-like long-term memory in large language models (LLMs) has been a central question for unlocking more general capabilities such as few-shot generalization. Existing memory frameworks and benchmarks focus on finding the optimal memory compression algorithm for higher performance in tasks that require recollection and sometimes further reasoning. However, such efforts have ended up building more human bias into the compression algorithm, through the search for the best prompts and memory architectures that suit specific benchmarks, rather than finding a general solution that would work on other data distributions. On the other hand, goal-directed search on uncompressed information could potentially exhibit superior performance because compression is lossy, and a predefined compression algorithm will not fit all raw data distributions. Here we present \textbf{SUMER} (\textbf{S}earch in \textbf{U}ncompressed \textbf{M}emory via \textbf{E}xperience \textbf{R}eplay), an end-to-end reinforcement learning agent with verifiable reward (RLVR) that learns to use search tools to gather information and answer a target question. On the LoCoMo dataset for long-context conversation understanding, SUMER with Qwen2.5-7B-Instruct learned to use search tools and outperformed all other biased memory compression approaches and also the full-context baseline, reaching SOTA performance (43\% gain over the prior best). We demonstrate that a simple search method applied to raw data outperforms goal-agnostic and biased compression algorithms in current long-context memory tasks, arguing for new paradigms and benchmarks that are more dynamic and autonomously scalable. Code for SUMER and all implemented baselines is publicly available at \url{https://github.com/zycyc/SUMER}.

\end{abstract}

\section{Introduction}
Large language models (LLMs) excel as general-purpose reasoners within a bounded context window, but their performance can degrade as inputs grow long or when relevant evidence is far from the query. Empirically, models often over-weight local or recent tokens and miss salient information in the middle of long inputs \citep{LiuLinHewittEtAl23}. Scaling context windows to hundreds of thousands or even millions of tokens (e.g., Gemini~1.5) improves access to evidence but does not fully resolve potential interference issues \citep{TeamGeorgievLeiEtAl24}. In more recent agentic settings, LLM agents often face issues when their context window is filled up fast and therefore require techniques to store and reuse information in an effective way.

There are two broad lines of work tackling long-horizon information use. The first seeks \emph{architectural} solutions: recurrence and state-space models (SSMs) that can carry state across unbounded sequences (e.g., RWKV, RetNet, Mamba) reduce quadratic attention costs and enable stateful processing that, in principle, extrapolates beyond a fixed window \citep{PengAlcaideAnthonyEtAl23,SunDongHuangEtAl23,GuDao24}. The second couples LLMs to \emph{external memory} or tools. Early neural memory systems (NTM, DNC, etc.) showed that differentiable read/write can extend algorithmic capabilities \citep{GravesWayneDanihelka14,GravesWayneReynoldsEtAl16}. In modern LLMs, retrieval‑augmented generation (RAG) conditions generation on retrieved context, with a design space spanning query rewriting, dense/sparse/hybrid retrieval, reranking, and compression \citep{GaoXiongGaoEtAl24}. Agentic tool use further allows models to plan multi‑step queries, browse, and call APIs (\emph{WebGPT} \citep{NakanoHiltonBalajiEtAl22}, \emph{Toolformer} \citep{SchickDwivedi-YuDessiEtAl23}, \emph{ReAct} \citep{YaoZhaoYuEtAl23}). Recent memory frameworks (\emph{MemGPT} \citep{PackerWoodersLinEtAl24}, \emph{A‑MEM} \citep{XuMeiGaoEtAl25}, \emph{Mem0} \citep{ChhikaraKhantAryanEtAl25}, etc.) focus on what to store and how to retrieve memories across long horizons.

Long‑context benchmarks (e.g., RULER \citep{HsiehSunKrimanEtAl24}, LongBench \citep{BaiLvZhangEtAl24}, and very long conversational memory in LoCoMo \citep{MaharanaLeeTulyakovEtAl24}) highlight that even with long windows or standard RAG pipelines, models still underperform on temporal, causal, and multi‑session reasoning. Many “memory” systems depend on \emph{goal‑agnostic} compression and fixed CRUD heuristics, which inject human design biases and could potentially discard details crucial for a downstream query unknown at compression time. By contrast, the “bitter lesson” in AI suggests that search and learning, rather than hand‑crafted knowledge representations, tend to win as tasks scale \citep{Sutton19}. In games, for example, search plus learned value/policy produces superhuman play that was impossible with human-designed playing algorithms (\emph{AlphaGo/Zero}) \citep{SilverHuangMaddisonEtAl16,SilverHubertSchrittwieserEtAl17}. For language agents, an analogous strategy is to defer compression and instead learn goal‑directed search over raw, uncompressed memory streams—only bringing in what is needed, when it is needed.

Here we present \textbf{SUMER} (\textbf{S}earch in \textbf{U}ncompressed \textbf{M}emory via \textbf{E}xperience \textbf{R}eplay): an end‑to‑end reinforcement learning (RL) agent with verifiable reward (RLVR) that learns to use simple tools to \emph{search}, \emph{inspect}, and \emph{answer} from raw conversational memory, rather than relying on pre‑defined compression. SUMER is trained with group relative policy optimization (GRPO) and multi‑turn masking so that tool responses are treated as context, while gradients flow only through the agent’s own actions and reasoning. Conceptually, our setup is complementary to recent RL‑for‑reasoning efforts (DeepSeek‑R1) and multi‑turn search‑RL frameworks (Search‑R1), but targets long‑context memory tasks where the evidence is distributed across many sessions; we adapt RLVR to reward only correct final answers while letting the agent discover effective search strategies \citep{ShaoWangZhuEtAl24,DeepSeek-AIGuoYangEtAl25,JinZengYueEtAl25}.

On LoCoMo’s long‑term conversational QA, SUMER starts from low zero‑shot accuracy, learns to chain search calls over training, and ultimately surpasses hand‑engineered memory baselines and full context, achieving new SOTA results with a net gain of \texttt{43}\% over prior best.

\begin{figure}[t]
  \centering
  \includegraphics[width=\linewidth]{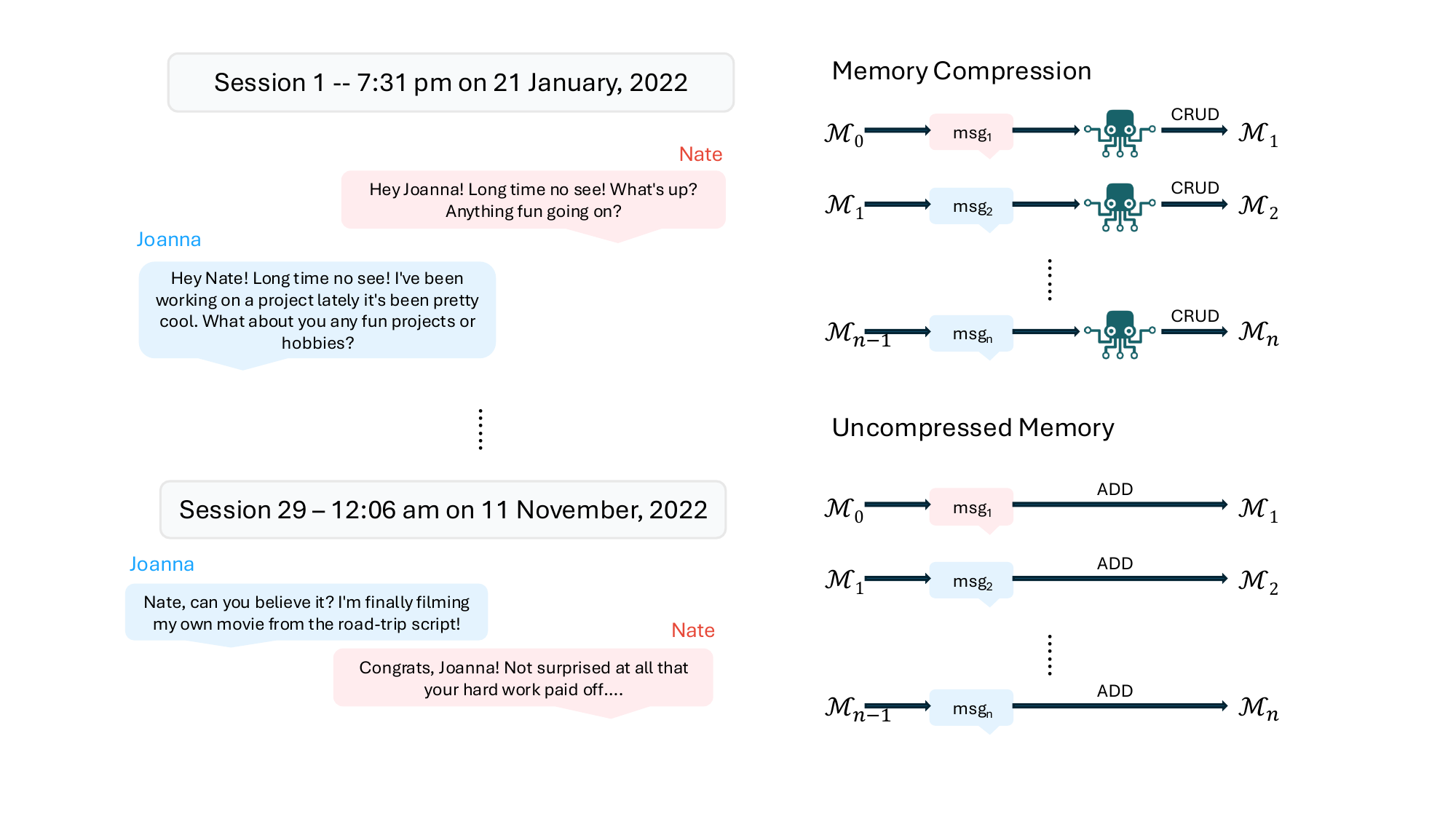}
  \caption{\textbf{Conversational memory vs.\ memory compression.}
  (left) Long-horizon chats span many sessions with distractors. 
  (right) Goal-agnostic memory compression applies Create, Read, Update, and Delete (CRUD) operations that can discard details later needed at query time, while our approach takes raw data as it is and directly adds it to the memory database for later search.
  }
  \label{fig:overview}
\end{figure}

\section{Related Works}
\subsection{External Memory in LLMs}
Neural networks paired with external memory have long been used to extend algorithmic and reasoning capabilities beyond what standard sequence models can manage, from Neural Turing Machines \citep{GravesWayneDanihelka14} and Memory Networks \citep{WestonChopraBordes15} to Differentiable Neural Computers \citep{GravesWayneReynoldsEtAl16} and many more \citep{BergesOguzHazizaEtAl24}. For large language models, the most popular method is retrieval-augmented generation (RAG), which retrieves task-relevant information from an external store using embedding vectors and conditions the model’s generation on it \citep{LewisPerezPiktusEtAl21}. Recent surveys organize RAG systems into stages covering query preparation, retrieval/indexing, and post-retrieval conditioning, and highlight design choices such as dense vs.\ sparse/hybrid retrieval, reranking, query rewriting, and compression/summarization for memory efficiency, etc. \citep{GaoXiongGaoEtAl24}.

One recent line of work on agentic LLMs aims to improve their performance on long tasks with many steps by implementing more sophisticated memory systems. \emph{MemGPT} introduces virtual context management and a tiered memory hierarchy, paging information in and out of the context window so the LLM operates over a manageable working set while keeping a persistent long-term state externally in disk memory \citep{PackerWoodersLinEtAl24}. \emph{A-MEM} proposes agentic memory that dynamically organizes experiences using Zettelkasten-like “notes,” explicit links between memories, and memory evolution (updates to prior notes when new information arrives), improving multi-hop and temporal retrieval for agents \citep{XuMeiGaoEtAl25}. \emph{Mem0} frames memory as production-ready infrastructure with LLM-driven ADD/UPDATE/DELETE/NOOP operations and reports strong long-horizon results, and graph-structured variants for relational reasoning \citep{ChhikaraKhantAryanEtAl25}, while Memory-R1 adds reinforcement learning (RL) of a memory manager and an answer agent to achieve stronger performance \citep{YanYangHuangEtAl25}; \emph{GraphRAG} builds knowledge graphs to support multi-hop discovery and retrieval beyond flat chunks \citep{EdgeTrinhChengEtAl25}; \emph{MemMachine} positions itself as a dedicated memory layer for agents, separating \emph{profile}, \emph{short-term} (episodic/summaries), and \emph{long-term} stores with reranking and consolidation \citep{MemMachine25}.

\subsection{RLVR and Multi‑Turn Agentic Tool Use}
RLVR refers to doing RL on tasks whose outcomes can be automatically checked by a program (and are therefore verifiable). For example, math problems with exact answers or code judged by test suites. Recent work shows that large‑scale RLVR can elicit stronger multi‑step reasoning than SFT alone, including GRPO objectives that stabilize training by normalizing rewards within sets of candidates \citep{ShaoWangZhuEtAl24}. Building on this, \citet{DeepSeek-AIGuoYangEtAl25} train reasoning models (R1/R1‑Zero) with RL to improve stepwise solutions, reporting emergent behaviors like self‑verification.

Early agentic methods taught models to interleave reasoning with actions (e.g., search APIs) within a single trajectory via prompting or supervised traces, for example, \emph{ReAct} couples thoughts and acts \citep{YaoZhaoYuEtAl23}, and \emph{WebGPT} uses browsing with imitation learning + preference optimization \citep{NakanoHiltonBalajiEtAl22}. Recent RLVR work extends this to \emph{multi‑turn} settings where the model repeatedly queries tools, reads results, and adapts its plan before committing to a final answer. \citet{JinZengYueEtAl25} (Search-R1) demonstrate end‑to‑end RL that teaches an LLM when to query a searcher and how to reason over retrieved evidence.

\subsection{Trainable Search over Memory}
For LLMs, test-time search improves reasoning and problem solving: \emph{Self-Consistency} searches over multiple chains-of-thought \citep{WangWeiSchuurmansEtAl23}; \emph{Tree-of-Thoughts} explores/prunes thought trees \citep{YaoYuZhaoEtAl23}; and \emph{DeepSWE} selects from sampled agentic coding trajectories for higher accuracy in coding tasks \citep{LuoJainSinghEtAl25}. With RLVR, a model can be trained to when and how to search according to a policy that is optimal for the target task. \emph{Search-R1} uses outcome rewards to teach when to search, what to query, and how to integrate results across multiple turns\citep{JinZengYueEtAl25}. Our setting mirrors this but targets long‑term conversational memory: instead of compress‑then‑retrieve, SUMER performs task‑conditioned search over uncompressed logs and optimizes the search policy for response accuracy. More recently, \emph{MEM1} learns to modify its memory bank to better perform tasks like Q\&A, outperforming methods that only compress memory, such as A-MEM \citep{ZhouQuWuEtAl25}. Similar to Search-R1, MEM1 targets knowledge corpuses like Wikipedia, and demonstrates the advantages of learned, goal-directed search over memory compression.

\begin{figure}[t]
  \centering
  \includegraphics[width=\linewidth]{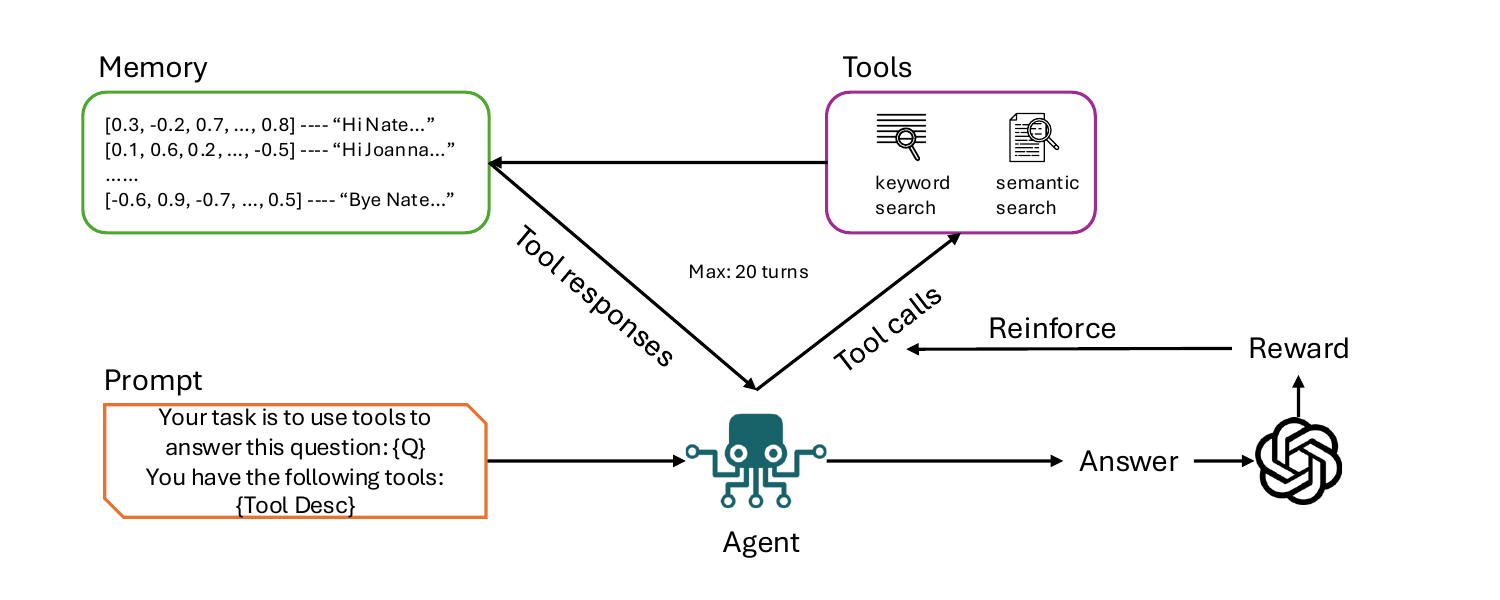}
  \caption{\textbf{SUMER training loop with tool use and RLVR.}
  The agent calls keyword/semantic search across multiple turns, then submits an answer for a verifiable reward. Retrieved tool tokens are visible as context but masked from the policy loss, so learning focuses on the agent’s outputs.}
  \label{fig:rl-loop}
\end{figure}

\section{Method}

We present SUMER, a RL framework that trains LLM agents to autonomously discover and utilize relevant memories through multi-turn tool interactions for question-answering tasks. Unlike traditional memory management approaches with explicit CRUD operations, SUMER empowers agents to learn effective memory search and retrieval strategies through outcome-driven rewards.

\subsection{Problem Formulation}

Consider a multi-session dialogue comprising $N$ sessions $\mathcal{S} = \{s_1, s_2, \ldots, s_N\}$, where each session $s_i$ contains multiple turns of conversation between two speakers. The objective is to answer questions $\mathcal{Q} = \{q_1, q_2, \ldots, q_M\}$ that require synthesizing information distributed across temporally distant sessions. We augment an LLM agent with an external memory bank $\mathcal{M}$ to maintain persistent knowledge beyond the model's limited context window.

Given a target question $q$ and a pre-populated memory bank $\mathcal{M}$, the agent engages in multi-turn interactions using two specialized tools: \texttt{search\_memory} for semantic/RAG-based and keyword-based retrieval and \texttt{submit\_answer} for final response generation. The challenge lies in learning effective search strategies and memory utilization patterns that maximize answer accuracy through RL.

\subsection{System Architecture}

SUMER employs a single LLM agent trained via GRPO \citep{ShaoWangZhuEtAl24} to interact with memory through a tool-augmented framework. The policy $\pi_\theta$ learns to search, retrieve, and reason over memories through multi-turn tool interactions. Given a question $q$ and access to memory tools $\mathcal{T}$, the agent generates a sequence of tool calls with tool responses following every tool call:
\begin{equation}
    (a_1, \ldots, a_T, y) \sim \pi_\theta(\cdot | q, \mathcal{T}, \mathcal{M})
\end{equation}
where $a_t$ represents tool calls at turn $t$, and $y$ is the final answer to the question.

During preprocessing, conversation-level memories (i.e., individual messages with metadata) are initialized from the LoCoMo dataset \citep{MaharanaLeeTulyakovEtAl24} and embedded using the Qwen3-Embedding-0.6B model \citep{ZhangLiLongEtAl25}, which generates 1024-dimensional dense vectors. These pre-computed embeddings were then added as-is to the Langmem memory bank to enable efficient semantic search during training. During memory search, to enrich the context of each memory, 2 messages right before and 2 messages right after the found memory were concatenated with the memory to form a memory group and fed to the agent. The \texttt{search\_memory} tool provides two search modes:

\textbf{Semantic Search}: Uses pre-computed Qwen3-Embedding-0.6B embeddings to find the $k$ most similar memories to a natural language query. The 1024-dimensional embeddings are compared using cosine similarity in the embedding space.

\textbf{Keyword Search}: Returns all memories where all specified keywords appear in either the content or metadata fields. This enables precise filtering when the agent knows specific terms to search for.

Both search modes support filtering by speaker and session, allowing the agent to narrow its search scope. The agent must learn through RL which search strategy and filters are most effective for different question types.

The generated answer is saved and compared with ground truth when \texttt{submit\_answer} is called, or when one the following conditions is met: 1) length of agent-environment interaction history exceeding context window of the LLM; 2) agent reaching maximum tool use turns (20); 3) no tool calls detected.

\subsection{Reinforcement Learning with GRPO}

We train the agent with GRPO \citep{ShaoWangZhuEtAl24}, which replaces critic-based advantages with \emph{group-normalized} rewards computed from $G$ sampled rollouts per prompt. For each question $q$ we draw a group of trajectories $\{\tau^{(i)}\}_{i=1}^{G}$ from the behavior policy $\pi_{\text{old}}$, obtain terminal rewards $\{r_i\}_{i=1}^{G}$, and compute a standardized advantage per rollout:
\begin{equation}
A_i \;=\; \frac{r_i - \mu_r}{\sigma_r + \epsilon}, 
\qquad
\mu_r \;=\; \frac{1}{G}\sum_{j=1}^{G} r_j,
\qquad
\sigma_r \;=\; \sqrt{\frac{1}{G}\sum_{j=1}^{G} (r_j - \mu_r)^2}.
\end{equation}
Here $i\in\{1,\dots,G\}$ indexes the rollouts in the group; $r_i$ is the scalar terminal reward; $\mu_r$ and $\sigma_r$ are the group mean and standard deviation; and $\epsilon>0$ is a small constant for numerical stability. In our experiments, we use $G{=}8$ trajectories per question and cap each trajectory at $20$ assistant turns \citep{JinZengYueEtAl25}.

\subsubsection{GRPO Objective}

Let $o_i=(o_{i,1},\dots,o_{i,|o_i|})$ denote the sequence of \emph{policy-generated} tokens for trajectory $\tau^{(i)}$ (tool outputs are not counted here). We define the token-level likelihood ratio
\begin{equation}
\rho_{i,t} \;=\; 
\frac{\pi_{\theta}\!\left(o_{i,t}\,\middle|\, q, o_{i,<t}\right)}
     {\pi_{\text{old}}\!\left(o_{i,t}\,\middle|\, q, o_{i,<t}\right)},
\end{equation}
where $\pi_{\theta}$ is the current policy, $\pi_{\text{old}}$ is the behavior policy, $o_{i,<t}$ is the token prefix, and $t$ indexes tokens. We broadcast the rollout-level advantage to tokens and apply loss masking via
\begin{equation}
\hat{A}_{i,t} \;=\; m_{i,t}\, A_i,
\end{equation}
where $m_{i,t}\in\{0,1\}$ masks out non-learned tokens (prompts and tool responses) by setting them to $0$ and keeps learnable tokens (agent outputs) at $1$.

Our training objective is based on GRPO \citep{ShaoWangZhuEtAl24,DeepSeek-AIGuoYangEtAl25}, with higher clipping, without any KL regularization term \citep{YuZhangZhuEtAl25}, and without entropy loss \citep{LuoJainSinghEtAl25}:
\begin{equation}
\label{eq:grpo}
J(\theta) \;=\;
\mathbb{E}\!\left[
\frac{1}{G}\sum_{i=1}^{G}\;
\sum_{t=1}^{|o_i|}
\min\!\Big(
\rho_{i,t}\,\hat{A}_{i,t},\;
\operatorname{clip}\!\left(\rho_{i,t},\, 1-\epsilon_{\text{low}},\, 1+\epsilon_{\text{high}}\right)\,\hat{A}_{i,t}
\Big)
\right].
\end{equation}

\subsubsection{Multi-Turn Tool Interactions and Masking}

Each trajectory $\tau^{(i)}$ consists of multiple assistant turns where the agent can invoke tools. The trajectory likelihood decomposes as:
\begin{equation}
    \pi_\theta(\tau | s_0) = \prod_{t=1}^{T} \pi_\theta(a_t | s_0, a_{1:t-1}, o_{1:t-1})
\end{equation}
where $a_t$ is the tool calls (text generation and tool calls) at turn $t$, and $o_{1:t-1}$ are the tool responses from previous turns. The agent can make up to 5 parallel tool calls per turn.

During training, we apply selective masking to focus learning on agent-generated content while providing tool responses as context. Specifically, we mask out prompts and tool responses and only train on the tokens that the agent generated. This masking ensures the agent learns to generate effective tool calls and reasoning while not focusing on predicting prompts or tool responses.

\subsection{Reward Function}

The agent is trained with a reward combining an LLM-judge correctness signal and an F1 score between the predicted and gold answers:

\begin{equation}
R =
\begin{cases}
\mathbb{J}\bigl(y_{\text{pred}}, y_{\text{gold}}\bigr)\, F_{1}\bigl(y_{\text{pred}}, y_{\text{gold}}\bigr), 
& \text{if answer submitted}, \\[4pt]
-1, 
& \text{if no answer submitted}
\end{cases}
\end{equation}

Correctness is evaluated using an LLM-as-judge approach with the \texttt{gpt-oss-120b} model \citep{OpenAIAgarwalAhmadEtAl25}. The judge checks semantic equivalence rather than exact string matching, allowing paraphrases as long as the answer is factually correct. To shape the output format, we additionally compute the token-level F1 score:

\begin{equation}
F_{1}(y_{\text{pred}}, y_{\text{gold}}) = \frac{2 \cdot \text{Precision} \cdot \text{Recall}}{\text{Precision} + \text{Recall}}
\end{equation}

where Precision and Recall are computed over the sets of tokens in the predicted and gold answers. This encourages the agent to match the style and brevity of the gold answers (typically short phrases) rather than producing overly long explanations that would still satisfy the LLM judge on semantics.

The LLM judge uses a structured prompt requesting binary classification (CORRECT/WRONG) in JSON format, with evaluation criteria emphasizing topical alignment and factual accuracy over exact wording. Importantly, only trajectories that successfully call \texttt{submit\_answer} receive non-zero rewards. This encourages the agent to learn when it has gathered sufficient information to answer, rather than searching indefinitely. Trajectories that exceed the turn limit without submitting an answer receive $R=-1$.

\subsection{Training and Validation Data}

Training uses 1 conversation (conv-48, first conversation after shuffling with a seed=42) out of the 10 conversations in LoCoMo \citep{MaharanaLeeTulyakovEtAl24}, with the 9 other conversations used as validation data. LoCoMo is a benchmark for evaluating very long-term conversational memory, consisting of 10 high-quality multi-session conversations generated through a machine-human pipeline and verified by human annotators for consistency and grounding.

Each conversation in LoCoMo contains an average of 27.2 sessions (ranging from 19 to 32 sessions), 588.2 turns (ranging from 369 to 689 turns), and approximately 17,390 tokens (ranging from 10,424 to 21,014 tokens).\footnote{Token counts are estimated using word count $\times$ 1.3 as an approximation. Exact counts may vary depending on the tokenizer used.} The training conversation (conv-48) specifically contains 30 sessions spanning 8 months (January to September 2023), with 681 dialogue turns totaling approximately 17,644 tokens. The conversation features two speakers (Deborah and Jolene) engaged in naturalistic long-term dialogue.

The benchmark includes 1,540 question-answer pairs across all 10 conversations (conv-48 contains 191 questions), categorized into four types: (1) \textit{single-hop questions} answerable from a single turn (54.6\% of questions), (2) \textit{multi-hop questions} requiring reasoning across multiple conversation turns (18.3\%), (3) \textit{open-domain questions} requiring inference beyond explicitly stated information (6.2\%), and (4) \textit{temporal questions} requiring temporal understanding of events (20.8\%). The dataset also contains adversarial questions (category 5), but these are excluded from evaluation as they lack ground truth labels.

Validation uses greedy decoding (sampling temperature 0) with a single trajectory per question. We evaluate every 50 steps.

\section{Experiments}

\subsection{Experiment Setup}

\subsubsection{Dataset} We performed minimal preprocessing to the LoCoMo \citep{MaharanaLeeTulyakovEtAl24} dataset to include timestamp of sessions and individual messages and then directly added each individual message as memory into the Langmem data structure (which could be replaced by a simple dictionary). We split the dataset to have 1 conversation as training data, and the other 9 conversations as validation data. We omitted the adversarial category due to lack of ground truth labels.

\subsubsection{Baselines} We obtained baseline code from the following repositories: Mem0 (\url{https://github.com/mem0ai/mem0/tree/main}, which has code for RAG, Full Context, Langmem, and Mem0), A-MEM (\url{https://github.com/WujiangXu/A-mem}), and MemMachine (\url{https://github.com/MemMachine/MemMachine}). We compare with the following baselines after adapting their code to use our local LLM setting and keeping other configurations the same (see Table~\ref{tab:sumer-config}): 1) \textbf{RAG}: We segmented the entire conversation into chunks of 500 tokens and retrieved the most relevant chunk using the target question as the query. We appended the retrieved text to the target question and obtained the answer from the LLM; 2) \textbf{Full Context}: Similar to RAG, except that instead of chunks of 500 tokens we used the entire conversation history as context; 3) \textbf{Langmem}: We used autonomous LangGraph agents for each speaker to manage their own memories. The agents processed conversations, autonomously stored relevant information in a local vector store, and independently searched memories to generate responses that were combined to obtain the final answer from the LLM. Because the Qwen-2.5-7B-Instruct model only has a context window of 32k, we had to cut the agent generation window down to 8k and left 24k for agent trajectory to run the eval script without error; 4) \textbf{A-Mem}: The system uses Zettelkasten-inspired memory organization with explicit links between memories, hybrid retrieval combining BM25 (Best Matching 25) and semantic search ($\alpha=0.5$ to balance between the two), and memory evolution enabled with a threshold of 100 retrievals (default in documentation). For each question, we used LLM-based keyword generation followed by top-$k$ ($k=10$) memory retrieval, then prompt the LLM to answer based on retrieved context. 5) \textbf{Mem0}: We extracted personalized memories from paired conversation messages for each speaker using Mem0's API as of October 29, 2025 (non-local LLM) and retrieved the top 30 most relevant memories per speaker using the target question as the query to obtain the answer from the local LLM; 6) \textbf{MemMachine}: We ingested each conversation into MemMachine's episodic memory system, where each message was stored as a memory episode with its speaker, timestamp, and metadata. During evaluation, for each target question, we queried the episodic memory system to retrieve up to 30 most relevant memory episodes. The retrieved episodes were provided as context to the LLM to answer the question.

\subsubsection{Implementation Details}

We implement SUMER using the VERL framework \citep{ShengZhangYeEtAl25} for distributed reinforcement learning. Our system runs on 8 NVIDIA H100 GPUs (80GB each) with tensor model parallelism (size=2) and Ulysses sequence parallelism (size=4) for efficient distributed training. We use Qwen-2.5-7B-Instruct as the base model with gradient checkpointing and FSDP offloading to manage memory constraints. The system supports prompts up to 8192 tokens and responses up to 24576 tokens with overflow filtering disabled to accommodate long multi-turn conversations. As a design choice, we used Qwen3-Embedding-0.6B instead of the more commonly used text-embedding-3-small for embedding, and gpt-oss-120b instead of gpt-4o-mini for LLM judge, which led to faster iteration of experiments.

During training, we use temperature $\tau=1.0$ for exploration with $G=8$ trajectories sampled per question, while validation employs greedy decoding ($\tau=0$) with a single trajectory per question for deterministic evaluation. Our batch configuration uses a global batch size of 32 with micro-batch size 4 per GPU and mini-batch size 32 for policy updates. We optimize with learning rate $1 \times 10^{-6}$ without KL regularization to focus purely on reward maximization. The distributed memory system utilizes GPU memory at 0.5 utilization for SGLang rollout workers across 32 agent workers. We validate every 50 steps and log 30 examples for qualitative analysis. For reward computation, we use gpt-oss-120b with temperature $\tau=0$ to ensure deterministic binary classification in our LLM-as-judge evaluation.

\begin{table*}[t]
  \centering
  \setlength{\tabcolsep}{3.2pt}
  \renewcommand{\arraystretch}{1.08}
  \begin{adjustbox}{width=\textwidth}
  \begin{tabular}{lrrrrrrrrrrrrrrr}
    \toprule
    \multicolumn{1}{l}{\textbf{Method}} &
    \multicolumn{3}{c}{\textbf{Single-Hop}} &
    \multicolumn{3}{c}{\textbf{Multi-Hop}} &
    \multicolumn{3}{c}{\makecell{\textbf{Open}\\\textbf{Domain}}} &
    \multicolumn{3}{c}{\textbf{Temporal}} &
    \multicolumn{3}{c}{\textbf{Overall}} \\
    \midrule
     & F1$\uparrow$ & B1$\uparrow$ & J$\uparrow$ & F1$\uparrow$ & B1$\uparrow$ & J$\uparrow$ & F1$\uparrow$ & B1$\uparrow$ & J$\uparrow$ & F1$\uparrow$ & B1$\uparrow$ & J$\uparrow$ & F1$\uparrow$ & B1$\uparrow$ & J$\uparrow$ \\
    \midrule
    RAG & 25.90 & 12.09 & 12.09 & 17.02 & 12.09 & 21.28 & 17.20 & 13.90 & 37.50 & 15.37 & 12.80 & 13.08 & 24.97 & 19.89 & 35.84 \\
    Full Context & 27.10 & 18.25 & 63.02 & 19.42 & 14.62 & 33.69 & 12.88 & 11.35 & \textbf{39.58} & 10.59 & 7.95 & 17.45 & 21.37 & 15.01 & 46.69 \\
    Langmem & 8.42 & 10.04 & 21.20 & 13.41 & 15.91 & 21.63 & 9.95 & 10.62 & 25.00 & 6.25 & 6.18 & 4.36 & 10.35 & 8.98 & 17.99 \\
    A-MEM & 35.36 & 30.46 & 46.70 & 20.54 & 13.85 & 24.11 & 11.91 & 10.62 & 27.34 & 31.34 & 26.32 & 25.23 & 27.28 & 23.13 & 32.00 \\
    Mem0 & 34.38 & 29.76 & 46.25 & 27.83 & 20.27 & 31.56 & 14.97 & 11.65 & 31.25 & 36.20 & 28.89 & 22.43 & 32.35 & 26.71 & 37.66 \\
    MemMachine & 48.18 & 41.78 & 44.35 & \textbf{32.86} & \textbf{23.18} & 24.82 & 14.76 & 11.20 & 19.79 & 37.60 & 28.85 & 17.76 & 41.09 & 33.77 & 33.70 \\
    \rowcolor{black!5} SUMER-Base & 34.98 & 30.30 & 64.45 & 16.90 & 13.36 & 36.78 & 13.48 & 11.53 & 36.05 & 25.10 & 21.22 & 22.22 & 28.07 & 23.95 & 48.55 \\
    \rowcolor{black!5} SUMER-GRPO & \textbf{61.82} & \textbf{56.55} & \textbf{79.53} & 28.45 & 21.85 & \textbf{44.83} & \textbf{19.98} & \textbf{17.45} & 39.53 & \textbf{42.23} & \textbf{37.66} & \textbf{62.72} & \textbf{48.65} & \textbf{43.44} & \textbf{66.79} \\
    \bottomrule
  \end{tabular}
  \end{adjustbox}
  \caption{Main results on LoCoMo validation (9 held-out conversations).}
  \label{tab:main}
\end{table*}

\subsection{Main Results}

Table~\ref{tab:main} compares SUMER (SUMER-Base: pre-RL, SUMER-GRPO: post-RL) against standard RAG, full-context prompting, and several goal-agnostic memory systems on LoCoMo, reporting token-level F1, BLEU-1 (B1), and LLM-judge correctness (J). Across all question types, SUMER trained with GRPO (SUMER-GRPO) achieves the best overall performance, nearly doubling judge accuracy relative to the strongest compression-based baseline. Specifically, compared to MemMachine, SUMER-GRPO improves overall F1 from 41.09 to 48.65 (+7.56), B1 from 33.77 to 43.44 (+9.67), and J from 33.70 to 66.79 (+33.09, \(\sim\)2$\times$). Even relative to our own pre-RL search baseline (SUMER-Base), GRPO training yields a substantial +18.24 gain in J (48.55 \(\rightarrow\) 66.79), corresponding to a 37.57\% relative improvement, along with consistent gains in F1 and B1.  

Decomposed by question type, SUMER-GRPO dominates or matches baselines everywhere. On single-hop questions, it attains 61.82 F1 / 56.55 B1 / 79.53 J, improving J by more than 15 points over the best non-RL variant. Multi-hop questions remain the most challenging regime: SUMER-GRPO improves J to 44.83, outperforming all baselines on judge accuracy while exhibiting a modest trade-off in F1 and B1 relative to MemMachine. Open-domain questions show smaller absolute gains, but SUMER-GRPO still matches or slightly exceeds the best prior J while delivering higher F1/B1 scores. For temporal reasoning, where correctly locating events in a long conversation is critical, SUMER-GRPO achieves 42.23 F1 / 37.66 B1 / 62.72 J, achieving large margins over all other baselines. 

\begin{figure}[ht]
  \centering
  \includegraphics[width=\linewidth]{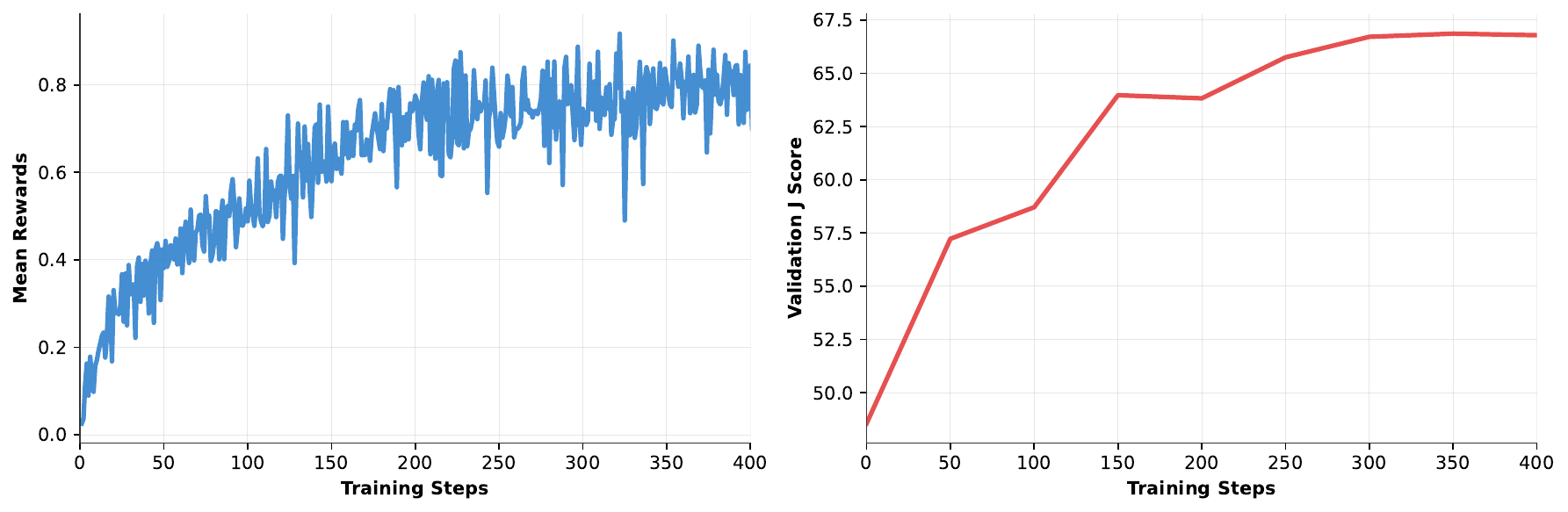}
  \caption{\textbf{SUMER training curves.} \textit{Left:} Mean rewards during training. \textit{Right:} Validation performance.}
  \label{fig:training-curves}
\end{figure}

Figure~\ref{fig:training-curves} illustrates that these improvements are realized through a stable RL learning process. The mean reward climbs steadily from around 0 to around 0.8 over 400 training steps, while validation J increases from 48.55 to 66.79. This indicates that SUMER learns more effective search strategies over time rather than simply overfitting to the training conversation. Overall, the results support our central claim: a goal-directed search policy over uncompressed conversational memory can outperform hand-engineered CRUD-style compression pipelines on long-context memory QA.

\subsection{Ablation Studies}

\begin{figure}[t]
  \centering
  \includegraphics[width=\linewidth]{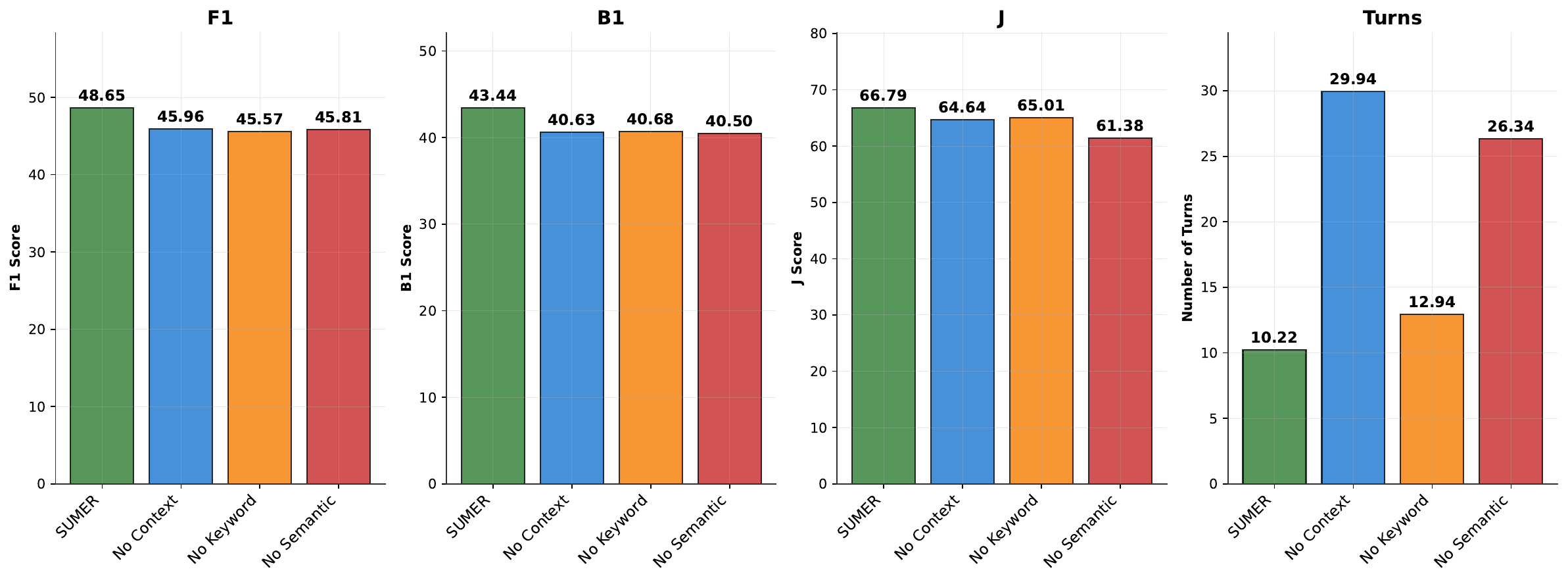}
  \caption{\textbf{Comparison between ablations.} SUMER outperformed all ablation conditions in F1, B1, J scores, and the number of turns to finish the task. Without temporal context of the retrieved memory, the no context variant requires more agentic search turns to gather enough information, and semantic search is more efficient than keyword search in general. }
  \label{fig:ablations}
\end{figure}

To understand which components of SUMER contribute most to performance, we conduct ablations that disable 1) temporal context around retrieved memories (No Context), 2) keyword-based search (No Keyword), and 3) semantic search over embeddings (No Semantic). Figure~\ref{fig:ablations} summarizes final validation performance and the average number of search turns, and Table~\ref{tab:ablations} (Appendix) reports initial, final, and peak metrics for each configuration.  

All ablated variants still benefit substantially from GRPO training. For example, No Context improves its overall J from 38.32 to 64.64 (+26.32, +68.67\% relative), No Semantic improves from 40.47 to 61.38 (+20.90, +51.65\%), and No Keyword improves from 49.52 to 65.01 (+15.49, +31.29\%). This indicates that RL is powerful enough to discover reasonably effective search strategies even when the toolset is partially crippled. 

Nonetheless, the full SUMER configuration consistently achieves the best trade-off between accuracy and efficiency. With all tools enabled, SUMER reaches 48.65 F1 / 43.44 B1 / 66.79 J while requiring only 10.22 tool-using turns on average. Removing temporal context around retrieved messages (No Context) leads to slightly lower final J (64.64) but almost triples the number of turns (29.94), suggesting that local temporal neighborhoods are important for quickly gathering sufficient evidence. Disabling semantic search (No Semantic) has the largest negative effect on J among the ablations (down to 61.38) and substantially increases the number of turns (26.34), implying that keyword search alone is a less efficient way to navigate long conversational histories. In contrast, removing keyword search (No Keyword) yields relatively mild degradation (65.01 J, 12.94 turns), indicating that semantic retrieval is still more efficient at finding the most semantically relevant information while keyword search provides complementary precision in a subset of cases.  

Taken together, these ablations support two conclusions. First, RL consistently improves performance across a range of configurations, showing that SUMER is robust to imperfections in the memory contents and tools provided. Second, the combination of semantic search, keyword search, and local temporal context yields both higher correctness and more efficient trajectories, confirming that more efficient search over richer raw information benefits current long-context memory tasks. Even without aggressive optimization, search over raw contents is still preferred to goal-agnostic compression in this setting.

\section{Discussion}

\subsection{Limitations and future directions}
This work is not primarily a proposal for a new search algorithm, but rather an argument about the relative value of search versus compression for long-context memory tasks. Our results suggest that when the end goal is accurate recovery of information from the original interaction history, aggressive compression of episodic memory may be counterproductive. Many current memory frameworks introduce additional inductive biases and discard information in ways that can help on narrowly defined benchmarks, but are not aligned with schema-based generalization in humans, where shared structures across diverse experiences are extracted to form abstractions.

From a broader perspective, a genuinely strong lifelong agent should be equipped with at least: 1) the ability to continually update an internal state of the world as new information arrives; 2) the ability to reliably access and reuse past experiences, so as not to repeat the same mistakes; 3) the ability to extract common patterns across experiences to support genuine generalization, akin to continued pretraining but with some control over what data to learn from. Existing long-context benchmarks such as LoCoMo do not meaningfully probe any of these capabilities. Instead, they largely resemble extended pattern-matching and question-answering setups similar in spirit to benchmarks before the LLM era. As a result, they may systematically under-estimate the importance of world-modeling and schema learning, and over-emphasize local retrieval over relatively short conversational horizons.

Our experimental setup also has practical limitations. Due to API and resource constraints, we were unable to train with the same GPT-4o-mini and \texttt{text-embedding-3-small} configurations used in prior work, and instead relied on Qwen-based models for both policy learning and retrieval. This mismatch makes it difficult to directly compare our absolute numbers to previously reported results. In addition, the LoCoMo dataset we study does not exceed the base model’s context window, so our experiments do not fully capture the regime where the conversation history is vastly longer than what can be naively fed into the model. In realistic long-term memory scenarios, quality of search and the biases in compression are likely to matter more differently.

Finally, the search policy we study is deliberately simple. The current paper does not introduce a sophisticated new agentic search method; rather, it demonstrates that a minimal agentic search procedure, when trained with RL, can already outperform SOTA compression-based memory frameworks and full context. Future work can move in two complementary directions. On the algorithmic side, more expressive search policies, richer tool use, and tighter integration between retrieval, world modeling, and planning may further improve performance. On the evaluation side, there is a clear need for more demanding benchmarks that require longer-term memory over histories far beyond the context window, and create conditions under which compression is plausibly beneficial (e.g., for distilling stable facts or schemas), while naive search alone may underperform. Such benchmarks would provide a more realistic testbed for studying the trade-offs between search, compression, and schema formation in agents.

\subsection{Conclusion}
We instantiated a simple search-based agent that operates directly over uncompressed conversational histories and trained it with RL to optimize downstream QA performance. On the LoCoMo benchmark, this agent substantially outperforms a range of compression-based memory systems, nearly doubling LLM-judge accuracy while using a relatively lightweight search procedure. These results show that, on current long-context benchmarks, a straightforward goal-directed search policy can be markedly more effective than carefully engineered compression pipelines.

\clearpage

\printbibliography

\clearpage
\appendix
\renewcommand\thesection{\Alph{section}}
\counterwithin{figure}{section}
\counterwithin{table}{section}
\renewcommand\thefigure{\thesection\arabic{figure}}
\renewcommand\thetable{\thesection\arabic{table}}

\section*{Appendix}
\section{Additional Quantitative Results}
\label{app:ablations}

In this section we report a consolidated summary of training dynamics for SUMER and the three ablation variants. Table~\ref{tab:ablations} lists initial and final validation performance, as well as absolute and relative improvements, for F1, BLEU-1 (B1), and LLM-judge correctness (J). These numbers complement the main results in Table~\ref{tab:main} by making explicit how much RL training improves each configuration over its own starting point.

\begin{table}[htbp]
\centering
\caption{Performance summary across all configurations. Values are averaged over the 9 LoCoMo validation conversations. $\Delta$ Abs and $\Delta$ Rel report absolute and relative improvements from the initial checkpoint.}
\label{tab:ablations}
\begin{tabular}{llrrrrr}
\toprule
Configuration & Metric & Initial & Final & $\Delta$ Abs & $\Delta$ Rel (\%) \\
\midrule
SUMER & F1 & 28.07 & 48.65 & +20.58 & +73.32 \\
SUMER & B1 & 23.95 & 43.44 & +19.49 & +81.37 \\
SUMER & J & 48.55 & 66.79 & +18.24 & +37.56 \\
No Context & F1 & 24.29 & 45.96 & +21.67 & +89.24 \\
No Context & B1 & 20.68 & 40.63 & +19.96 & +96.51 \\
No Context & J & 38.32 & 64.64 & +26.32 & +68.67 \\
No Keyword & F1 & 30.06 & 45.57 & +15.51 & +51.60 \\
No Keyword & B1 & 25.66 & 40.68 & +15.02 & +58.52 \\
No Keyword & J & 49.52 & 65.01 & +15.49 & +31.29 \\
No Semantic & F1 & 23.54 & 45.81 & +22.27 & +94.60 \\
No Semantic & B1 & 19.97 & 40.50 & +20.53 & +102.83 \\
No Semantic & J & 40.47 & 61.38 & +20.90 & +51.65 \\
\bottomrule
\end{tabular}
\end{table}

\section{Prompt Templates and Evaluation Setup}
\label{app:prompts}

This section provides the exact text prompts used to train and evaluate SUMER. Figure~\ref{fig:system_prompt} shows the high-level system prompt that instructs the agent to treat the task as memory search followed by answer submission. Figure~\ref{fig:training_prompt} gives the full training prompt template, including the description of the memory database and the initial context shown to the agent for each question. Figure~\ref{fig:judge_prompt} shows the LLM-as-judge prompt used by the gpt-oss-120b model to compute binary correctness labels that feed into the reward.

\begin{figure}[htbp!]
\centering
\fbox{\begin{minipage}{0.9\textwidth}
\small
\textbf{System Prompt:}

You are an expert at searching memory databases for question-answering. Your goal is to search through an existing memory database to find relevant information and provide an answer to a target question. Available tools are described below.
\end{minipage}}
\caption{SUMER agent system prompt used during training to guide memory search and answer submission behavior.}
\label{fig:system_prompt}
\end{figure}

\begin{figure}[htbp!]
\centering
\fbox{\begin{minipage}{0.9\textwidth}
\small
\textbf{Training Prompt Template:}

You have access to the following memory database: \{total\_memories\} total memories (\{breakdown\_str\}) across \{num\_sessions\} sessions between \{speakers\}. The database contains \{level\_descriptor\}. Each memory includes speaker, session, timestamp, and source metadata.

\textbf{Target Question: \{question\}}

Your task is to search through the memory database to find relevant information that helps answer the above question, then submit your final answer.

\textbf{Instructions:}
\begin{enumerate}
    \item Use the search\_memory tool to find relevant memories that could help answer the question
    \item You may search multiple times with different queries and search types to gather comprehensive information
    \item Once you have found sufficient information, use the submit\_answer tool to provide your final answer
\end{enumerate}

\textbf{IMPORTANT}: The question may require:
\begin{itemize}
    \item Information from a single session
    \item Synthesizing information from multiple sessions
    \item Temporal reasoning across conversations
    \item Integrating speaker information with general knowledge
\end{itemize}

\textbf{INSTRUCTIONS for answering the question:}
\begin{enumerate}
    \item Carefully analyze all provided memories
    \item Pay special attention to any timestamps or temporal information
    \item If the question asks about a specific event or fact, look for direct evidence in the memories
    \item If the memories contain contradictory information, prioritize the most recent information
    \item If there is a question about time references (like "last year", "two months ago", etc.), calculate the actual date based on context
    \item Always convert relative time references to specific dates, months, or years when possible
    \item Focus only on the content of the memories provided
    \item The answer should be concise and direct, less than 5-6 words when possible
\end{enumerate}

You have up to 20 turns to search for information and submit your answer. Focus on finding the most relevant memories to answer the question accurately.

Here is some relevant context from the conversation database that may help answer the question:

======\\
Conversation Memories - Speaker 1\\
======

-- Memory 1 --

\{timestamp1\}

\{memory1\}

...

-- Memory 5 --

\{timestamp5\}

\{memory5\}

======\\
Conversation Memories - Speaker 2\\
======

-- Memory 1 --

\{timestamp1\}

\{memory1\}

...

-- Memory 5 --

\{timestamp5\}

\{memory5\}

Now, please search for more specific information and submit your final answer using the submit\_answer tool.

\end{minipage}}
\caption{Complete training prompt template. Each training sample receives this prompt, including a high-level description of the memory database and a small set of example memories, before the agent begins multi-turn search.}
\label{fig:training_prompt}
\end{figure}

\begin{figure}[htbp!]
\centering
\fbox{\begin{minipage}{0.9\textwidth}
\small
\textbf{LLM-as-Judge Binary Classification Prompt:}

Your task is to label an answer to a question as ’CORRECT’ or ’WRONG’. You will be given the following data:

    (1) a question (posed by one user to another user),

    (2) a ’gold’ (ground truth) answer,

    (3) a generated answer

which you will score as CORRECT/WRONG.

The point of the question is to ask about something one user should know about the other user based on their prior conversations.
The gold answer will usually be a concise and short answer that includes the referenced topic, for example:

Question: Do you remember what I got the last time I went to Hawaii?

Gold answer: A shell necklace

The generated answer might be much longer, but you should be generous with your grading - as long as it touches on the same topic as the gold answer, it should be counted as CORRECT.

For time related questions, the gold answer will be a specific date, month, year, etc. The generated answer might be much longer or use relative time references (like "last Tuesday" or "next month"), but you should be generous with your grading - as long as it refers to the same date or time period as the gold answer, it should be counted as CORRECT. Even if the format differs (e.g., "May 7th" vs "7 May"), consider it CORRECT if it's the same date.

Now it's time for the real question:

Question: \{question\}

Gold answer: \{golden\_answer\}

Generated answer: \{generated\_answer\}

First, provide a short (one sentence) explanation of your reasoning, then finish with CORRECT or WRONG.

Do NOT include both CORRECT and WRONG in your response, or it will break the evaluation script.

Just return the label CORRECT or WRONG in a json format with the key as "label".
\end{minipage}}
\caption{Binary classification prompt for the gpt-oss-120b judge model. The reward system uses this prompt for binary correctness evaluation, employing a generous evaluation strategy that accepts semantically correct answers even when surface form differs.}
\label{fig:judge_prompt}
\end{figure}

\section{Memory Search Tool Behavior}
\label{app:tool}

For completeness, we include an example output from the \texttt{search\_memory} tool in Figure \ref{fig:search_response}. This illustrates the format in which retrieved memories are returned to the agent, including timestamps, speaker information, surrounding context messages, and a counter for the remaining tool turns. Both keyword-based and semantic search modes use this format; the only difference is how the relevant memories are selected.

\begin{figure}[htbp!]
\centering
\fbox{\begin{minipage}{0.9\textwidth}
\small
\textbf{Search Memory Tool Response:}

Found 8 relevant memories using \texttt{keyword\_search} (filtered by: \texttt{speaker: Nate}):

Memory 1 [Time: 11:54 am on 2 May 2022]:

Nate: Gaming has been my focus - practicing a lot and even winning a few tournaments. Last week I won my second tournament!

Joanna: Wow, congrats! What game were you playing?

Nate: Thanks! I usually play CS:GO, but I tried my hand at the local Street Fighter tournament this time since I play that game a lot with my friends, and turns out I'm really good!

Joanna: Nice! That must have been a surprise. How did it feel to finally win one?

Nate: It was super awesome! So much adrenaline went into that last match, and the other finalist even shook my hand! Enough about me though, how about you? What have you been up to?

Memory 2 [Time: 3:00 pm on 25 May 2022]:

Nate: Hey Jo! Been ages since we last talked. Here's something cool that happened the other day - I took Max for a walk and ran into this super nice couple who had a dog. It turns out they live close by. We decided to do doggy playdates, which is awesome considering we all need friends for our pets.

Joanna: Hey Nate! Great to hear from you. Sounds like a nice encounter on your walk. Connecting with others who have pets can be uplifting and rewarding.

Nate: It's like fate. Having a walking buddy forMax will be great. He really likes the other dog too!

...

[turns remaining: 17]

\end{minipage}}
\caption{Example output of the \texttt{search\_memory} tool using a \texttt{keyword\_search} query filtered by speaker. For each retrieved memory (a single message), the tool also surfaces up to two preceding and two subsequent messages for additional context when metadata is compatible.}
\label{fig:search_response}
\end{figure}

\section{Training Configuration}
\label{app:config}

Table~\ref{tab:sumer-config} lists the full set of hyperparameters and system settings used in the main SUMER experiment. This includes model choices, GRPO training settings, rollout limits, and hardware configuration. The main run used 400 GRPO steps on a single 8$\times$H100 node for policy training, with the judge model served on a separate node.

\begin{table}[htbp!]
\centering
\begin{tabular}{|l|l|}
\hline
\textbf{Parameter} & \textbf{Value} \\
\hline
Model & Qwen-2.5-7B-Instruct \\
Embedding Model & Qwen3-Embedding-0.6B \\
Judge Model & gpt-oss-120b \\
Training Algorithm & GRPO \\
Training Steps & 400 \\
Batch Size & 32 \\
Rollout Trajectories (G) & 8 \\
Max Assistant Turns & 20 \\
Max Parallel Tool Calls & 5 \\
Training Temperature & 1.0 \\
Validation Temperature & 0.0 \\
Learning Rate & $1 \times 10^{-6}$ \\
Clip Ratio High & 0.28 \\
Context Length & 8192 tokens \\
Response Length & 24576 tokens \\
Memory Shards & 32 Ray Actors \\
Hardware & 8x NVIDIA H100 (80GB) \\
Model Parallelism & Tensor=2, Sequence=4 \\
GPU Memory Utilization & 0.5 \\
Validation Frequency & Every 50 steps \\
\hline
\end{tabular}
\caption{Complete SUMER training configuration. The main SUMER experiment took around 21 hours on a node with 8 H100 GPUs, with the judge model served on a separate node.}
\label{tab:sumer-config}
\end{table}

\section{LoCoMo Category Definitions and Mapping}
\label{app:locomo}

For reproducibility, we document how we handled LoCoMo question categories. Table~\ref{tab:locomo_categories} summarizes the four non-adversarial categories used in our experiments, which correspond to the breakdown in Table~\ref{tab:main}. Adversarial questions were ignored due to missing ground-truth answers.

In addition, we found a mismatch between the human-readable descriptions in the LoCoMo white paper and the actual category IDs used in the released code. Table~\ref{tab:category_mapping} shows the mapping we used. All reported per-category scores in the main text are based on the \emph{source code} mapping.

\begin{table}[htbp!]
\centering
\begin{tabular}{|l|l|l|}
\hline
\textbf{Category} & \textbf{Code} & \textbf{Description} \\
\hline
Multi-hop & 1 & Requires reasoning across facts \\
Temporal & 2 & Time-sensitive questions \\
Open-domain & 3 & General knowledge questions \\
Single-hop & 4 & Direct factual questions \\
\hline
\end{tabular}
\caption{LoCoMo dataset question categorization used in our experiments. We ignored the adversarial category (code 5) due to lack of ground-truth labels.}
\label{tab:locomo_categories}
\end{table}

\begin{table}[htbp!]
\centering
\begin{tabular}{c|l|l}
\hline
\textbf{Category ID (JSON/Code)} & \textbf{White Paper (sequential description)} & \textbf{Source Code (actual mapping)} \\ \hline
1 & Single-hop & Multi-Hop \\
2 & Multi-Hop & Temporal Reasoning \\
3 & Temporal Reasoning & Open-Domain \\
4 & Open-Domain & Single-Hop \\
5 & Adversarial & Adversarial \\ \hline
\end{tabular}
\caption{Mapping between category IDs, white paper descriptions, and source code labels for LoCoMo. We follow the source-code mapping when computing per-category metrics.}
\label{tab:category_mapping}
\end{table}

\end{document}